\documentclass[a4paper, conference]{IEEEtran}
\IEEEoverridecommandlockouts
\usepackage[left=19.1mm, right=13.1mm, top=36.7mm, bottom=19.1mm]{geometry}
\usepackage{cite}
\usepackage{amsmath,amssymb,amsfonts}
\usepackage{graphicx}
\usepackage{textcomp}
\usepackage{xcolor}
\usepackage{algorithm}
\usepackage{amsmath}
\usepackage{amssymb}
\usepackage{algpseudocode} 
\usepackage{caption}
\usepackage{subcaption}

\usepackage{multirow}
\usepackage{tabularx}
\usepackage{longtable}
\usepackage{booktabs}

\DeclareMathOperator*{\argmin}{argmin}

\def\BibTeX{{\rm B\kern-.05em{\sc i\kern-.025em b}\kern-.08em
    T\kern-.1667em\lower.7ex\hbox{E}\kern-.125emX}}
\begin{document}

\title{\vspace{6.3mm}S3M: Semantic Segmentation Sparse Mapping for UAVs with RGB-D Camera\\

}
\author{\IEEEauthorblockN{Thanh Nguyen Canh\textsuperscript{1,3}, Van-Truong Nguyen\textsuperscript{2}, Xiem HoangVan\textsuperscript{1}, Armagan Elibol\textsuperscript{3}, Nak Young Chong\textsuperscript{3}
}
\IEEEauthorblockA{\textsuperscript{1}University of Engineering and Technology, Vietnam National University \\
Hanoi, Vietnam (\{canhthanh, xiemhoang\}@vnu.edu.vn)
\\
\textsuperscript{2}Department of Mechatronics Engineering, Hanoi University of Industry \\ Hanoi 159999, Vietnam (nguyenvantruong@haui.edu.vn)
\\
\textsuperscript{3}School of Information Science, Japan Advanced Institute of Science and Technology \\
Nomi, Ishikawa 923-1292, Japan (\{aelibol, nakyoung\}@jaist.ac.jp)
}
}
\maketitle

\begin{abstract}
Unmanned Aerial Vehicles (UAVs) hold immense potential for critical applications, such as search and rescue operations, where accurate perception of indoor environments is paramount. However, the concurrent amalgamation of localization, 3D reconstruction, and semantic segmentation presents a notable hurdle, especially in the context of UAVs equipped with constrained power and computational resources. This paper presents a novel approach to address challenges in semantic information extraction and utilization within UAV operations. Our system integrates state-of-the-art visual SLAM to estimate a comprehensive 6-DoF pose and advanced object segmentation methods at the back end. To improve the computational and storage efficiency of the framework, we adopt a streamlined voxel-based 3D map representation - OctoMap to build a working system. Furthermore, the fusion algorithm is incorporated to obtain the semantic information of each frame from the front-end SLAM task, and the corresponding point. By leveraging semantic information, our framework enhances the UAV's ability to perceive and navigate through indoor spaces, addressing challenges in pose estimation accuracy and uncertainty reduction. Through Gazebo simulations, we validate the efficacy of our proposed system and successfully embed our approach into a Jetson Xavier AGX unit for real-world applications.

\end{abstract}


\section{Introduction} \label{sec:introduction}
Over the past decades, UAVs have significantly impacted geoinformation acquisition in areas like firefight rescue, inspection, and agriculture.  Understanding the environment is crucial for advancing autonomous capabilities, including the UAV's self-location awareness and semantic 3D map creation. Semantic mapping, which combines environmental geometry estimation with semantic labeling, goes beyond traditional geometric mapping, enhancing UAVs' situational understanding and interactions. For instance, in the rescue mission, a robot relying solely on a traditional SLAM-generated map encounters challenges in performing complex tasks, such as: "maneuvering around the desk to locate a victim beside the bed". Besides that, this task is still challenging due to: (1) the inaccuracy of GPS indoors, (2) the cluttered environments, (3) the real-time process demands, and (4) the complexity of semantic maps. 

To address the aforementioned challenges, the adoption of Simultaneous Localization and Mapping (SLAM) techniques, especially Visual SLAM~\cite{campos2021orb, engel2014lsd, engel2017direct} emerges as a compelling solution in the realm of drone applications, which is characterized by its compact design and cost-effectiveness, offering a wealth of information to comprehend the field of view. Various algorithms have been developed for this task such as KinectFusion~\cite{izadi2011kinectfusion}, RGBD\_SLAM~\cite{endres2012evaluation}, ORB-SLAM~\cite{mur2015orb}, PTAM~\cite{klein2007parallel}, DSO\_SLAM~\cite{engel2017direct}, LSD\_SLAM~\cite{engel2014lsd}, and SVO\_SLAM~\cite{forster2014svo}. Their effectiveness however varies depending on each scenario and the environments in which the robot operates such as localization, mapping, and real-time processes. Additionally, the research on the fusion of the semantic segmentation CNNs with the visual SLAM has been investigated with some notable works, including Semantic Fusion~\cite{mccormac2017semanticfusion}, Mask Fusion~\cite{runz2018maskfusion}, SCFusion~\cite{wu2020scfusion}, Co-Fusion~\cite{runz2017co}, and DS-SLAM~\cite{yu2018ds}. Despite these advancements, achieving semantic reconstruction for UAVs remains challenging. Hence, our research aims to establish mapping with semantic data, vital for enabling UAVs to perform advanced autonomous tasks.
\begin{figure*}
    \centering
    \includegraphics[width=\linewidth]{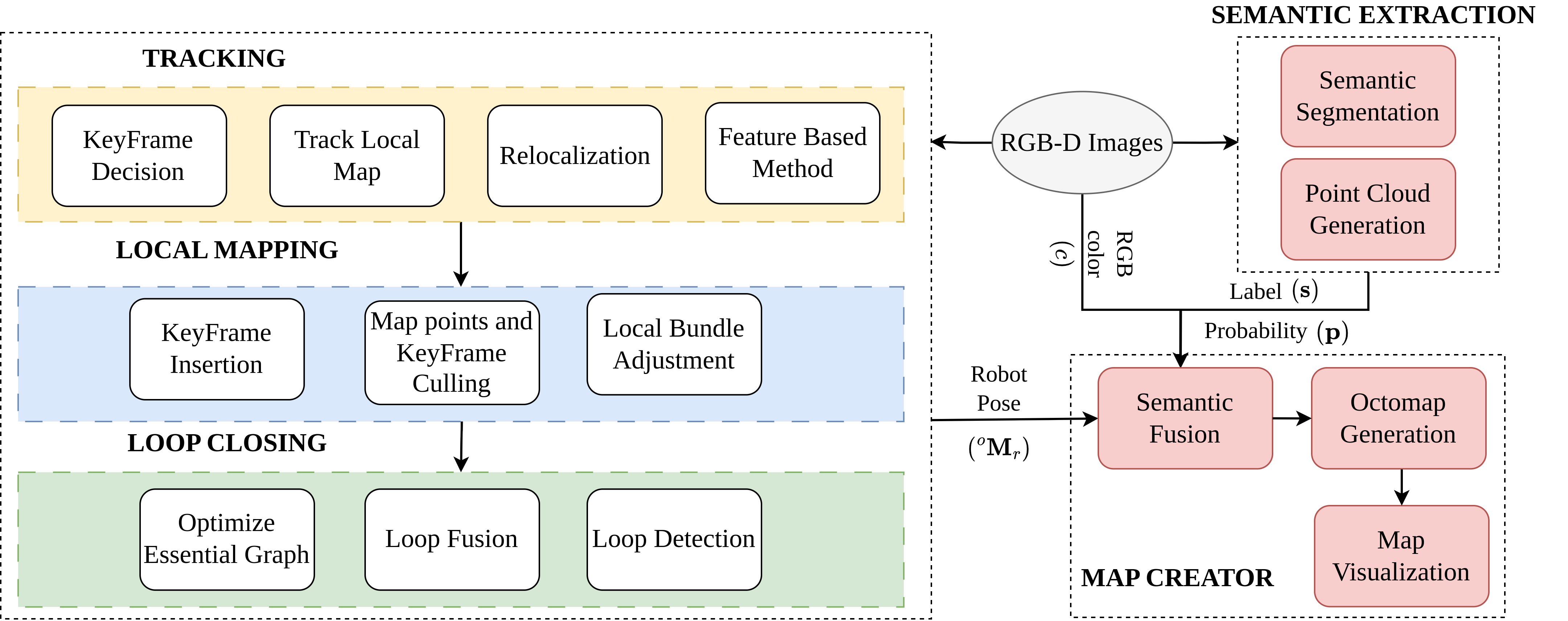}
    \caption{\textbf{Proposed S3M SLAM Architecture}: The system is composed of three units: a full 6 DoF pose estimation of the drone through ORB-SLAM3 (Tracking part - yellow, Local Mapping part - blue, Loop Closing part - green), a 3D semantic segmentation branch, and a semantic fusion scheme}
    \label{fig:proposed_system}
\end{figure*}
In this paper, we proposed an efficient Semantic Segmentation Sparse Mapping (S3M) SLAM system for incrementally constructing an object-level map using a localized RGB-D camera. The proposed system is organized into two main components: an RGB-D SLAM framework based on propagation utilizing Visual Odometry (VO) estimation and object instance segmentation-based semantic sparse map creation. To summarize, the main contributions of this work can be summarized as follows: \begin{itemize}
     \item A S3M SLAM system that has faster fully 6-DoF pose tracking and the capability to construct a semantic sparse map based on object segmentation information.
     \item A semantic fusion strategy based on geometric and semantic descriptions to incrementally update objects.
     \item An efficient representation and storage method using OctoMap of the front-end system, a memory-efficient alternative to point cloud data.
     \item The demonstrated capability of constructing semantically sparse maps in real-time on a compact, computation-limited platform via experiments on the Jetson Xavier AGX embedded computer.
 \end{itemize}
 
 The remainder of this paper is organized as follows: Sec \ref{sec:method} describing the proposed system based on RGB-D SLAM and object segmentation. The experiments conducted and results analysis are presented in Sec \ref{sec:exp}. Finally, Sec \ref{sec:conclusion} draws a conclusion with future works.
 
\section{Methodology} \label{sec:method}
The proposed S3M SLAM pipeline is illustrated in Fig.~\ref{fig:proposed_system}, which takes RGB-D sequences as input and progressively constructs a volumetric map enhanced with object instances. To achieve this, the RGB-D images undergo initial processing via a UAV pose tracking framework (Section~\ref{sec:posetracking}). Subsequently, an object instance segmentation method is applied to detect and extract semantic 3D objects from individual frames (Section~\ref{sec:semantic_segmen}). These identified objects are then integrated into a volumetric mapping framework to generate a dense map at the object level (Section~\ref{sec:semantic_fusion}). To enhance map quality, Octomap is utilized for noise removal and voxel grid downsampling to save space, with optimization for improved visual representation (Section~\ref{sec:map_creation}).
The system is implemented within a ROS framework, the widely used platform in the robotics community, the system leverages open-source tools, libraries, and conversions to simplify the development of intricate and robust robotics behaviors.
\subsection{Pose estimation} \label{sec:posetracking}
The accurate estimation of the UAV's pose is a critical step in our S3M SLAM pipeline. We employ the ORB-SLAM3 algorithm~\cite{campos2021orb} for robust and real-time camera pose estimation from RGB-D images. ORB-SLAM3 utilizes a monocular camera model and extends it to support stereo and RGB-D setups, making it well-suited for our UAV's sensor configuration. It encompasses three parallel threads: (1) Tracking, (2) Local Mapping, and (3) Loop Closing \cite{Zhang2018}. The pose estimation problem involves determining the position $(x, y, z)$ and orientation $(\phi, \theta, \psi)$ of the UAV in a global coordinate system. ORB-SLAM3 solves this problem by tracking a set of distinctive features in consecutive frames and establishing the correspondences between them. The estimated pose is obtained by minimizing the reprojection error between the observed feature locations and their predicted locations in the camera frame. Mathematically, given a set of $N$ observed 2D feature points $\mathbf{p_i}$ in the current RGB-D frame and their corresponding 3D points $\mathbf{P_i}$ in the world frame, the estimated camera pose \par
$^{\bar{o}} \mathbf{M}_c = \begin{bmatrix}
^{\bar{o}} \mathbf{R}_{c} & ^{\bar{o}} \mathbf{T}_{c} \\
0_{1 \times 3} & 1
 \end{bmatrix}= \begin{bmatrix}
^{\bar{o}} r_{c_{00}} & ^{\bar{o}} r_{c_{01}} & ^{\bar{o}} r_{c_{02}} & ^{\bar{o}} t_{c_{00}}\\ 
^{\bar{o}} r_{c_{10}} & ^{\bar{o}} r_{c_{11}} & ^{\bar{o}} r_{c_{12}} & ^{\bar{o}} t_{c_{10}}\\ 
^{\bar{o}} r_{c_{22}} & ^{\bar{o}} r_{c_{22}} & ^{\bar{o}} r_{c_{22}} & ^{\bar{o}} t_{c_{20}}\\ 
 0& 0 & 0 & 1
\end{bmatrix}$ \par can be obtained by solving the optimization problem:
\begin{equation} \label{eq:camera_pose}
    ^{\bar{o}} \mathbf{M}_c = \argmin_{^{\bar{o}} \mathbf{M}_c}\sum_{i=1}^N||\mathbf{p_i} - \pi (^{\bar{o}} \mathbf{M}_c \times \mathbf{P_i})||^2
\end{equation}

where $^{\bar{o}} \mathbf{R}_c, ^{\bar{o}} \mathbf{T}_c, ^{\bar{o}} \mathbf{M}_{c}$ are the rotation matrix, the translation matrix, and the transformation matrix between the world's coordinate frame ($\mathbf{\bar{O}}_{xyz}$) and the camera’s coordinate frame in ORB-SLAM3, respectively. $\pi(\cdot)$ is the projection function from 3D to 2D points, and $||\cdot||$ represents the Euclidean distance.

Since camera odometry obtained from Eq.~\ref{eq:camera_pose} and robot odometry have distinct world coordinates, we performed a calibration process. Denote $^o \mathbf{M}_r, ^o \mathbf{M}_c $ respectively as the transformation matrix representing the robot pose and camera pose relative to the robot's world frame ($\mathbf{O}_{xyz}$). The transformation $^o \mathbf{M}_c $ is then computed as:
\begin{equation}
    ^o \mathbf{M}_c = \begin{bmatrix}
^{\bar{o}} r_{c_{00}} & ^{\bar{o}} r_{c_{02}} & ^{\bar{o}} r_{c_{21}} & ^{\bar{o}} t_{c_{20}}\\ 
^{\bar{o}} r_{c_{20}} & ^{\bar{o}} r_{c_{11}} & ^{\bar{o}} r_{c_{01}} & ^{\bar{o}} t_{c_{00}}\\ 
^{\bar{o}} r_{c_{12}} & ^{\bar{o}} r_{c_{10}} & ^{\bar{o}} r_{c_{22}} & ^{\bar{o}} t_{c_{10}}\\ 
 0& 0 & 0 & 1
 \end{bmatrix}
\end{equation}

Let $^c \mathbf{M}_r$ represent the transformation matrix between the camera and the UAV's frame. The UAV's pose is given by:
\begin{equation}
    ^o \mathbf{M}_r = ^o \mathbf{M}_c  \times ^c \mathbf{M}_r
\end{equation}

% ORB-SLAM3 utilizes a bundle adjustment technique to refine the estimated pose and optimize the trajectory by minimizing the reprojection error over all the tracked feature points. This iterative optimization process ensures accurate and consistent pose estimates, which are essential for building an accurate semantic map. The robot pose estimation step provides the foundation for subsequent semantic mapping and object detection processes in our pipeline.
\begin{figure}
    \centering
    \includegraphics[width=\linewidth]{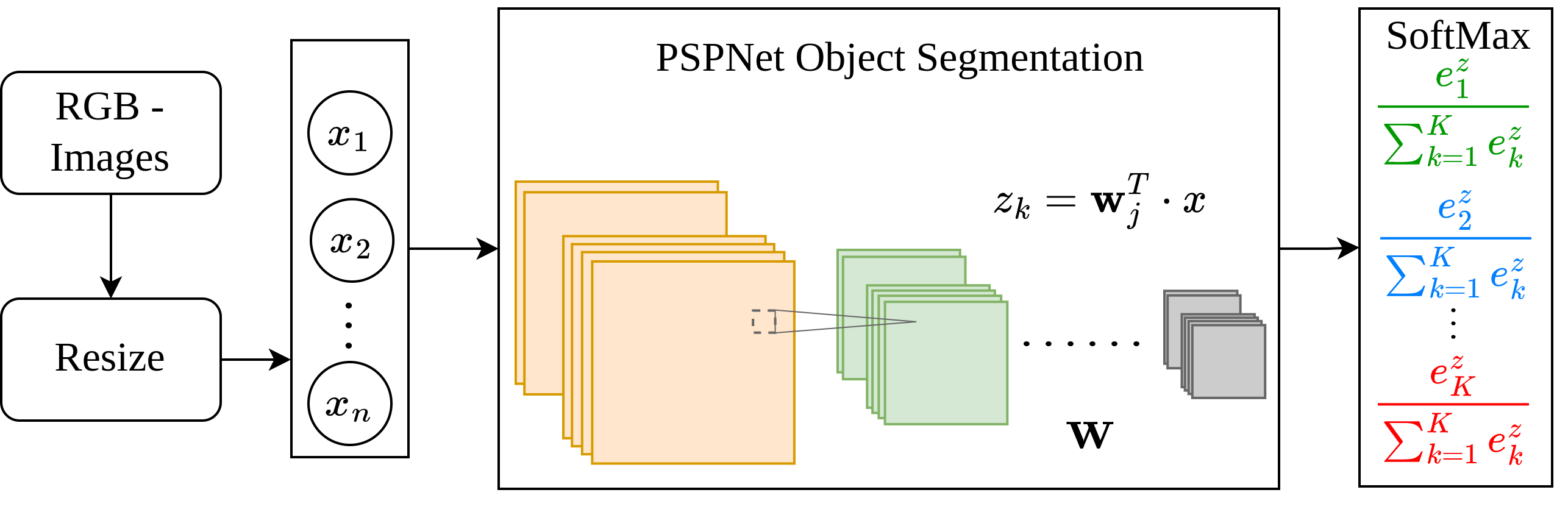}
    \caption{Structure of semantic segmentation model}
    \label{fig:segment}
\end{figure}
\subsection{Semantic segmentation} \label{sec:semantic_segmen}
In our methodology, semantic segmentation plays a vital role in extracting meaningful object instances from RGB-D images. This process is illustrated in Fig.~\ref{fig:segment}. Firstly, a color image is resized to the input size of the network. For our semantic segmentation network, we adopted the Pyramid Scene Parsing Network (PSPNet)~\cite{zhao2017pyramid} due to its proven effectiveness in generating accurate pixel-level semantic labels. PSPNet employs a multi-step process involving feature extraction with ResNet, pyramid pooling, convolutions on pooled feature maps, fusion of feature maps, and final convolutions to generate a class score map. This map assigns probabilities $p_i$ to pixels, enabling precise identification of object instances and their semantic labels. To finalize the process, a softmax activation is applied to the class score map, producing a probability distribution. Each pixel, along with its probability, is then selected and fused with the point cloud's pose. 
\subsection{Semanic fusion} \label{sec:semantic_fusion}
To achieve comprehensive scene understanding, it becomes imperative to integrate semantic labels across multiple views with translations. In addition to position and RGB data, semantics information is encoded within a point cloud. We represent this information using the vector $\mathbf{Q} = \begin{bmatrix}
\mathbf{t} & c &\mathbf{s} & \mathbf{p}
\end{bmatrix} ^ T$, where $\mathbf{t} \in \mathbb{R}^{3}$ and $c \in \mathbb{R}^{1}$ denote the 3D position and RGB color of a point cloud. This information is derived from RGB images and depth images. Furthermore, $\mathbf{s} \in \mathbb{R}^{k}$ and $\mathbf{p} \in \mathbb{R}^{k}$ symbolize $k$ semantic colors with the highest probability and their respective confidence scores associated with a point cloud. In each observation $O_i$, we calculate the probability for each semantic color within a given semantic set. Subsequently, the point featuring the highest probability is selected as the final decision. This approach ensures that semantic information is effectively incorporated into the point cloud, enabling a richer and more nuanced understanding of the scene across multiple viewpoints and translations. Algorithm~\ref{alg:SemanticFusion} outlines the semantic fusion process.
\begin{algorithm}
\caption{Semantic Segmentaion Fusion Approach}\label{alg:SemanticFusion}
\begin{algorithmic}[1]
\Require $\ \mathbf{Q}_1$  \Comment{Point cloud in Oservation 1}

         $\mathbf{Q}_2$  \Comment{Point cloud in Oservation 2}

         $\alpha$ \Comment{Trade of coefficient}
\Ensure $\mathbf{Q}_{fusion}$ 

\If{$\mathbf{Q}_1.\mathbf{s} = \mathbf{Q}_2.\mathbf{s}$}
    \State $\mathbf{Q}_{fusion} = \mathbf{Q}_1$
\Else

    \Comment{Probability for other unknown colors}
    \State $\bar{p_1} = 1 - \sum(\mathbf{Q}_1.\mathbf{p})$
    \State $\bar{p_2} = 1 - \sum(\mathbf{Q}_2.\mathbf{p})$
    
    \Comment{Synchronize data from $\mathbf{Q}_1$ to $\mathbf{Q}_2$}
    \For {each $label$ in $\mathbf{Q}_1.\mathbf{s}$ not in $\mathbf{Q}_2.\mathbf{s}$} 
        \State $(\mathbf{Q}_2.\mathbf{s}).push\_back(label)$
        
        \State $(\mathbf{Q}_2.\mathbf{p}).push\_back(\alpha \times \bar{p_2})$

        \State $\bar{p_2} = 1 - \sum(\mathbf{Q}_2.\mathbf{p})$
    \EndFor

    \Comment{Synchronize data from $\mathbf{Q}_2$ to $\mathbf{Q}_1$}
    \For {each $label$ in $\mathbf{Q}_2.\mathbf{s}$ not in $\mathbf{Q}_1.\mathbf{s}$} 
        \State $(\mathbf{Q}_1.\mathbf{s}).push\_back(label)$
        
        \State $(\mathbf{Q}_1.\mathbf{p}).push\_back(\alpha \times \bar{p_1})$

        \State $\bar{p_1} = 1 - \sum(\mathbf{Q}_1.\mathbf{p})$
    \EndFor

    \State $\mathbf{Q}_{fusion} = \mathbf{Q}_1$

    \Comment{Nomalize to probability distribution}
    \State $\mathbf{Q}_{fusion}.\mathbf{p} = \big (\mathbf{Q}_{1}.\mathbf{p} \times \mathbf{Q}_{2}.\mathbf{p} \big ) \big / \big (\sum(\mathbf{Q}_{1}.\mathbf{p} \times \mathbf{Q}_{2}.\mathbf{p} \big )$
\EndIf
\end{algorithmic}
\end{algorithm}

\subsection{Semantic map creation} \label{sec:map_creation}
In our approach, each keyframe retains the 3D point clouds, while the segmented 3D point clouds are preserved in alignment with the respective objects. However, point cloud-based maps often demand substantial storage space, rendering them unsuitable for modeling large-scale environments with limited memory and lack of structures to efficiently store each point, hindering search operations. 
% Moreover,  they lack structures to efficiently store each point, hindering search operations, and failing to provide volume information for individual points, limiting their usefulness for advanced tasks like path planning or grasp point selection.
To address these challenges, we adopted OctoMap~\cite{hornung2013octomap}, a probabilistic 3D mapping framework based on octrees. OctoMap presents a more efficient solution for storing occupancy status compared to point cloud maps, significantly reducing storage demands. Leveraging octrees, OctoMap divides spaces into small cubes, further subdivided into eight smaller cubes. Leaf nodes represent the smallest voxels, and a probabilistic model tackles issues like noise and range measurement errors by assigning probabilities to occupied or free states. This makes OctoMap an ideal choice for creating maps in our system, as it overcomes the limitations posed by traditional point cloud-based approaches. When a new 3D point is inserted, the log odds value for the voxel $i$ at time $t$ $(L(i|Z_{1:t-1})$ is computed using the log odds value accumulated up to time $t-1$ $(L(i|Z_{1:t-1})$:
\begin{equation}
    L(i|Z_{1:t}) = L(i|Z_{1:t-1}) + L(i|Z_{t})
\end{equation}
where,
\begin{equation}
    L(i) = \log \Big [ \frac{p(i)}{1-p(i)} \Big]
\end{equation}

Here, $Z_t$ represents the observed for a voxel at time $t$. $p(i)$ is the probability that the voxel $i$ contains an object or obstacle.

% The formula illustrates that the log-odds value of a voxel increases with repeated observations of occupancy, and decreases otherwise. Voxel occupancy is considered valid and reflected in the OctoMap only when its log-odds value exceeds a predefined threshold.

\section{Experimental Results} \label{sec:exp}

\subsection{UAVs Simulation}
\begin{figure}
    \centering
    \includegraphics[width=0.9\linewidth]{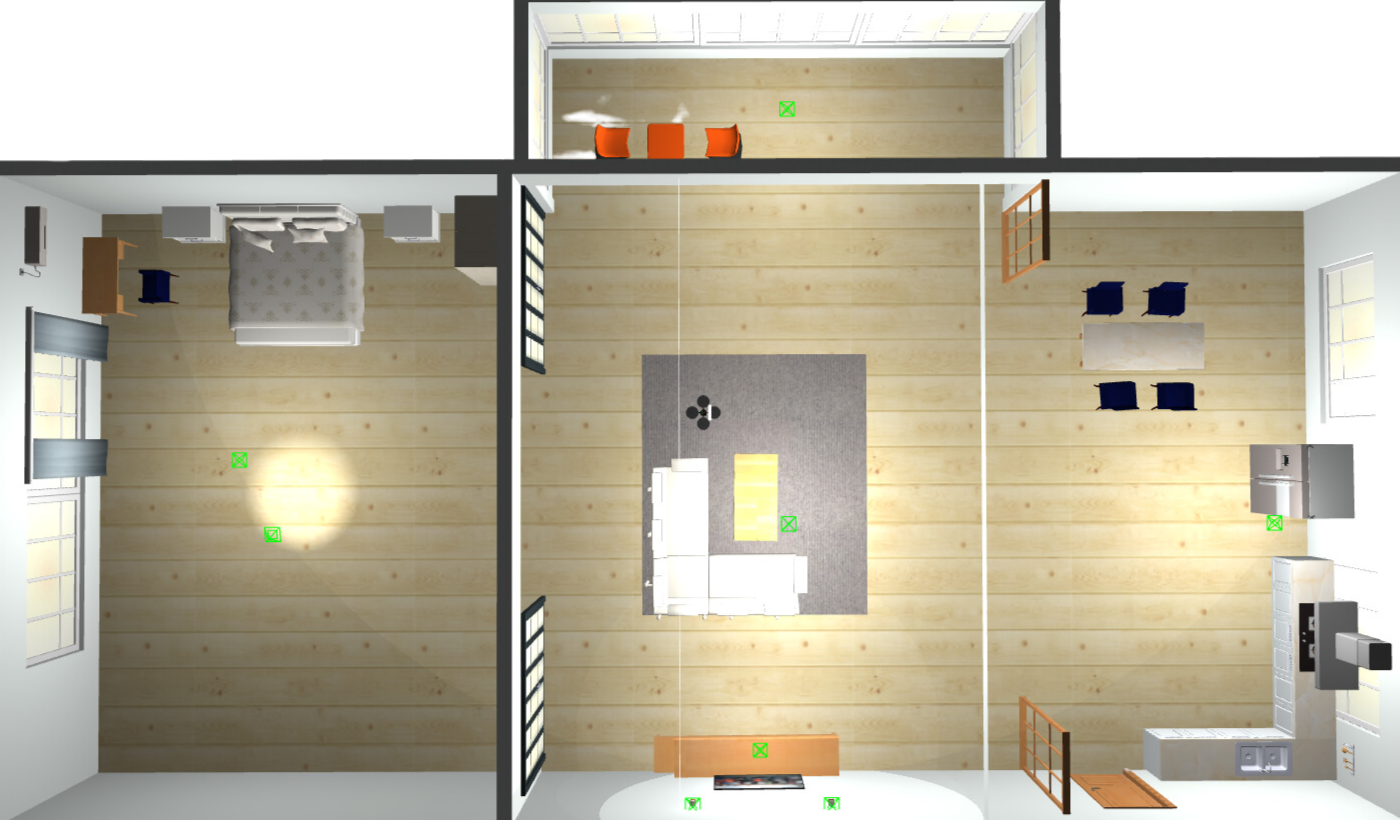}
    \caption{UAV and gazebo environment simulation}
    \label{fig:uav}
\end{figure}

The experimental evaluation of our proposed S3M SLAM system was conducted on the Hummingbird UAV platform equipped with a RealSense camera as shown in Fig.~\ref{fig:uav}. The Hummingbird UAV~\cite{wall2002hummingbird} is characterized by its lightweight design, enabling agile flight maneuvers and precise navigation in dynamic and challenging environments, such as those encountered in search and rescue operations. It is equipped with state-of-the-art flight control algorithms, ensuring stable and controlled flight behavior during the experiments. The RealSense D455 camera complements the UAV's capabilities by providing RGB-D data, which is crucial for accurate pose estimation and semantic information extraction.
\begin{figure}[!ht]
    \centering
    \begin{subfigure}[b]{0.24\textwidth}
    \centering
    \includegraphics[width=\textwidth]{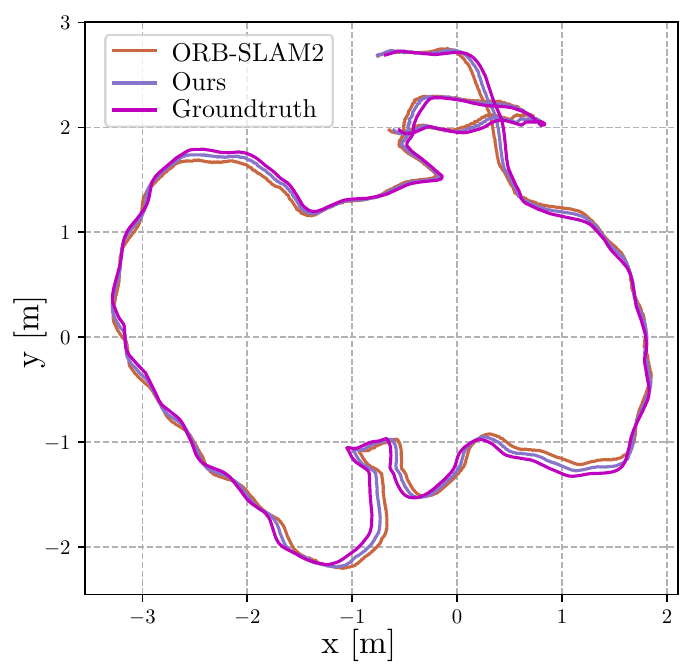}
    \caption{TUM Dataset}
    \label{fig:tumxy}
    \end{subfigure}
    % \hspace{0.5 cm}
    \centering
    \begin{subfigure}[b]{0.24\textwidth}
    \centering
    \includegraphics[width=\textwidth]{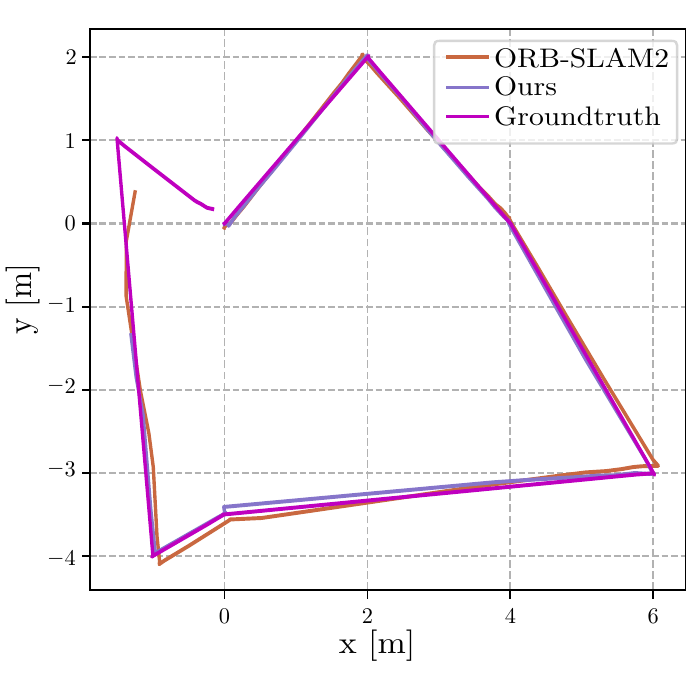}
    \caption{Gazebo Dataset }
    \label{fig:gazeboxy}
    \end{subfigure}
    
    \caption{The comparison of trajectory for ORB-SLAM2, Our system and ground truth in X-Y axis}
    \label{fig:xyestimate}
\end{figure}

\begin{figure}[!ht]
\centering
    \begin{subfigure}[b]{0.49\textwidth}
    \centering
    \includegraphics[width=\textwidth]{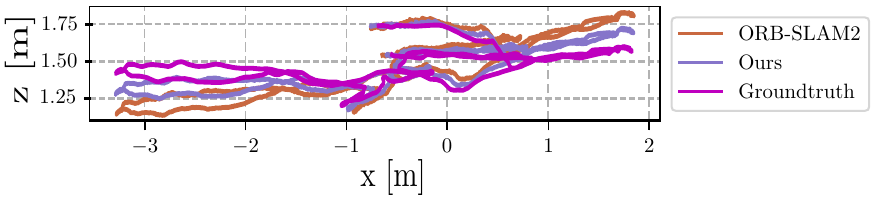}
    \caption{TUM Dataset}
    \label{fig:tumxz}
    \end{subfigure}
    \begin{subfigure}[b]{0.49\textwidth}
    \centering
    \includegraphics[width=\textwidth]{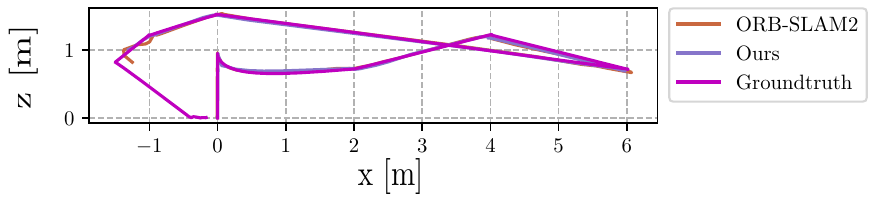}
    \caption{Gazebo Dataset}
    \label{fig:gazeboxz}
    \end{subfigure}
    
    \caption{The comparison of trajectory for ORB-SLAM2, our sytem and ground truth in X-Z axis}
    \label{fig:xzestimate}
\end{figure}

\subsection{Pose estimation evaluation}
\begin{figure}[!ht]
    \centering
    \begin{subfigure}[b]{0.49\textwidth}
    \centering
    \includegraphics[width=\textwidth]{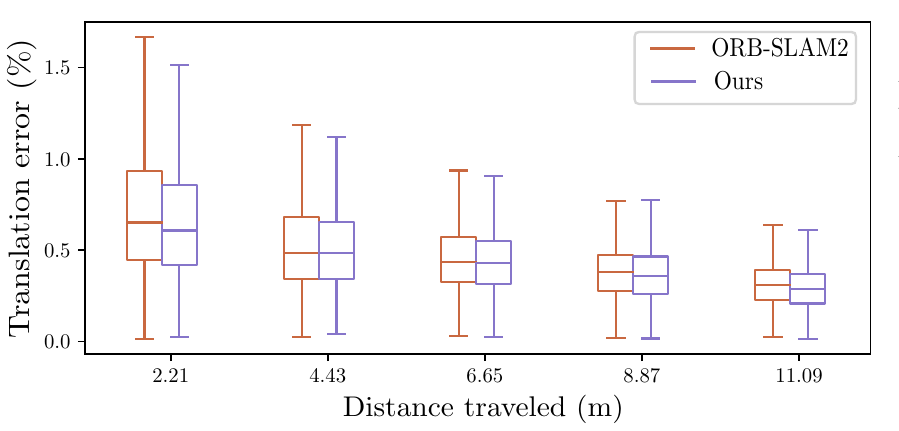}
    \caption{Translation Error in TUM dataset}
    \label{fig:tumrpetrans}
    \end{subfigure}
    % \hspace{0.5 cm}
    \centering
    \begin{subfigure}[b]{0.49\textwidth}
    \centering
    \includegraphics[width=\textwidth]{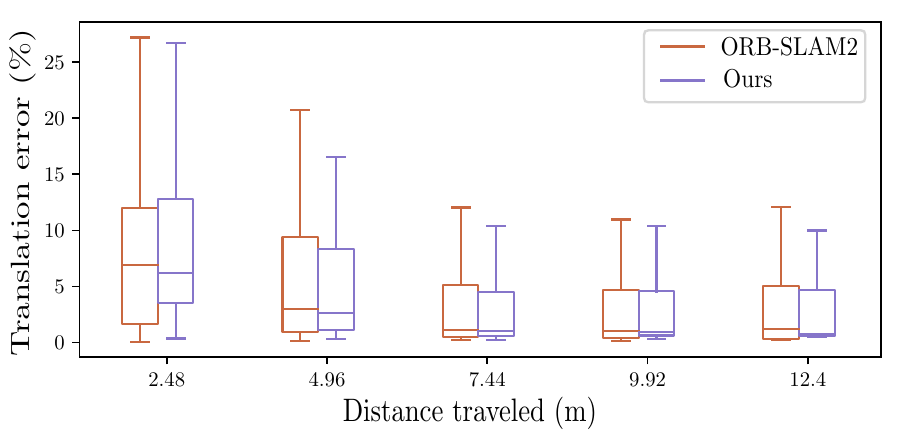}
    \caption{Translation Error in Gazebo dataset}
    \label{fig:gazeborpetrans}
    \end{subfigure}
    \centering
    \begin{subfigure}[b]{0.49\textwidth}
    \centering
    \includegraphics[width=\textwidth]{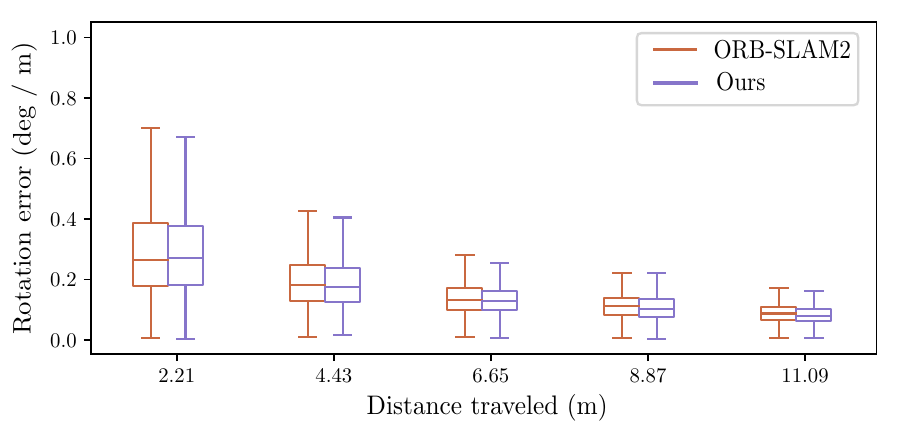}
    \caption{Rotation Error in TUM dataset}
    \label{fig:tumrperot}
    \end{subfigure}
    % \hspace{0.5 cm}
    \centering
    \begin{subfigure}[b]{0.49\textwidth}
    \centering
    \includegraphics[width=\textwidth]{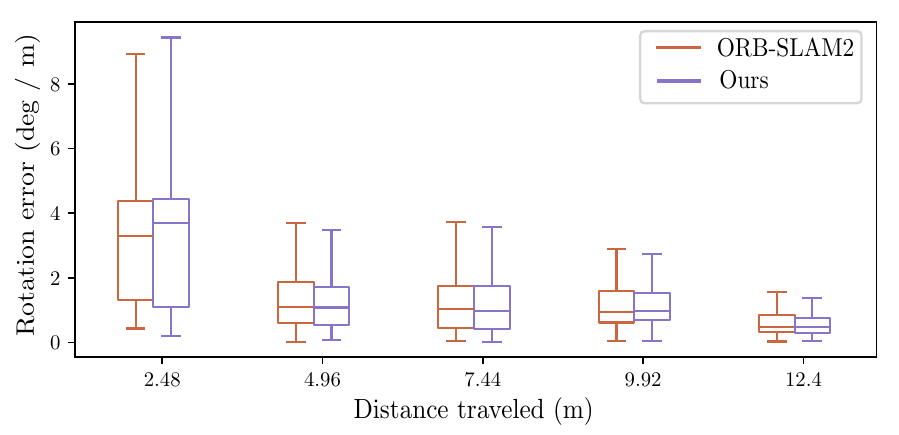}
    \caption{Rotation Error in Gazebo dataset}
    \label{fig:gazeborperot}
    \end{subfigure}
    
    \caption{Comparison of Relative Rose Error (RPE) between ORB-SLAM2 and Our system}
    \label{fig:rpeerror}
\end{figure}

\begin{figure}[!ht] 
    \centering
    \begin{subfigure}[b]{0.49\textwidth}
    \centering
    \includegraphics[width=\textwidth]{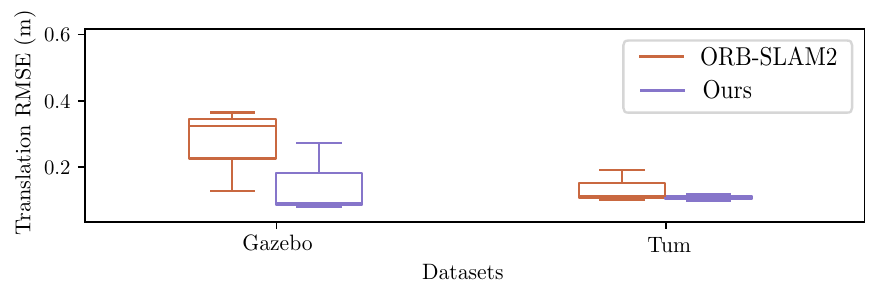}
    \caption{Translation Error}
    \label{fig:ate_trans}
    \end{subfigure}
    \centering
    \begin{subfigure}[b]{0.49\textwidth}
    \centering
    \includegraphics[width=\textwidth]{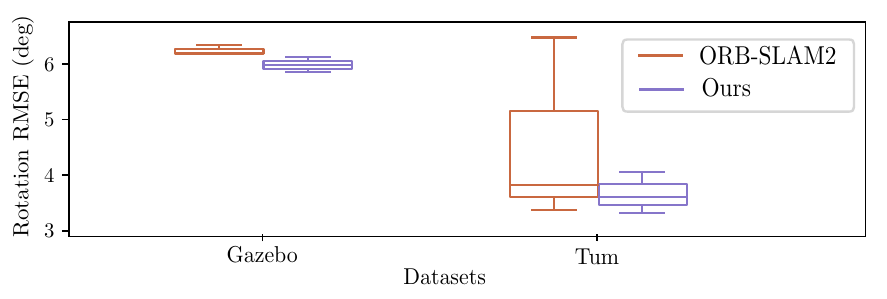}
    \caption{Rotation Error}
    \label{fig:ate_rot}
    \end{subfigure}
    
    \caption{The comparison of ORB-SLAM2 and Our system based on the RMSE of ATE}
    \label{fig:ate_error}
\end{figure}

To evaluate the pose estimation accuracy of the S3M system, we conducted evaluations on two distinct types of datasets: 1) TUM publicly available RGB-D data sequences ~\cite{brostow2009semantic}, 2) the simulation dataset obtained from Gazebo. The precision of 6-DoF pose estimation was evaluated using the Root Mean Square Error (RMSE) of Absolute Trajectory Error (ATE) and Relative Pose Error (RPE). Figs.~\ref{fig:xyestimate} and \ref{fig:xzestimate} show experiment results for pose trajectories, comparing ORB-SLAM2, our proposed system, and the ground truth across both datasets. Both ORB-SLAM2 and our system demonstrated accurate pose estimation and smooth movement within the environments, as depicted in Fig.~\ref{fig:rpeerror}. Finally, Fig.~\ref{fig:ate_error} presents the RMSE of ATE for all frames of both frameworks, reaffirming our system still ensures pose estimation performance.

\subsection{Training and evaluation on SUNRGBD dataset}
\begin{figure}
    \centering
    \includegraphics[width=\linewidth]{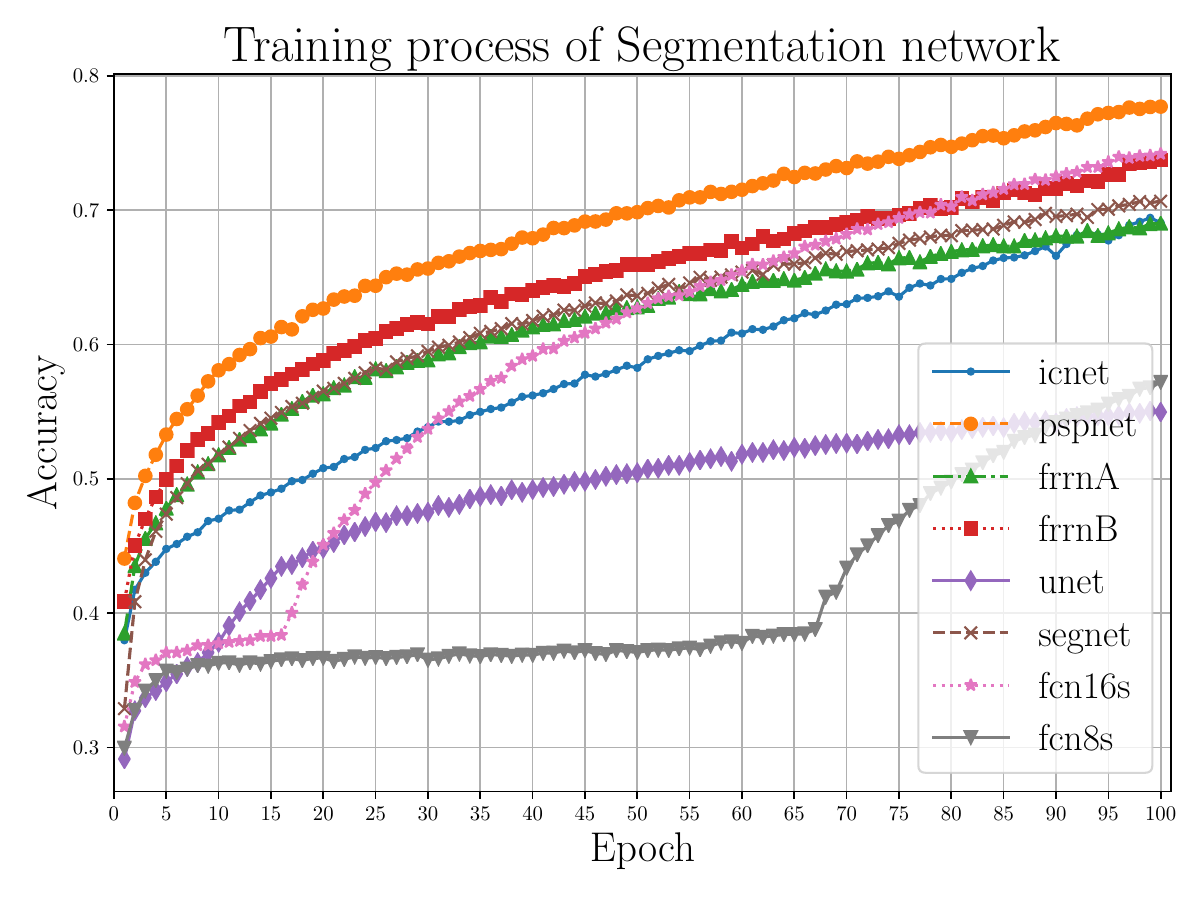}
    \caption{Training Models Assessment}
    \label{fig:seg_acc}
\end{figure}
SUNRGBD~\cite{song2015sun} stands as a widely adopted benchmark for evaluating semantic scene understanding. This dataset encompasses a total of $10,335$ images distributed across 38 semantic classes, with $5,285$ images earmarked for training and $5,050$ for validation. We selected 6 types of networks for training PSPNet~\cite{zhao2017pyramid}, ICNet~\cite{li2020icnet}, SegNet~\cite{badrinarayanan2017segnet}, UNet~\cite{ronneberger2015u}, FRRNs~\cite{pohlen2017full}, and FCNs~\cite{long2015fully}. In the case of FCNs, two network variants, denoted as 8s, and 16s, were utilized. Similarly, FRRNs were employed in both A and B settings. Each model underwent training with a maximum of 100 epochs and batch size of 2 on an Nvidia T4, and the best-performing model was chosen. For optimization, standard stochastic gradient descent was utilized, featuring a weight decay of 1e-3, a momentum of 0.9, and a learning rate of 0.01. The experiment results for each model tested in SUNRGBD are depicted in Fig. \ref{fig:seg_acc}. Among the models, PSPNet exhibited superior accuracy performance, prompting its selection as the segmentation model for integration into our system.
 
\subsection{Semantic Map}
\begin{figure*}[!ht] 
    \centering
    \begin{subfigure}[b]{0.3\textwidth}
    \centering
    \includegraphics[width=\textwidth]{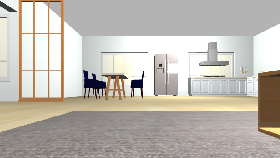}
    \end{subfigure}
    \hspace{0.1 cm}
    \begin{subfigure}[b]{0.3\textwidth} 
    \centering
    \includegraphics[width=\textwidth]{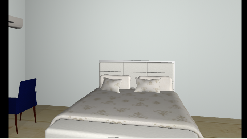}
    \end{subfigure}
    \hspace{0.1 cm}
    \begin{subfigure}[b]{0.3\textwidth}
    \centering
    \includegraphics[width=\textwidth]{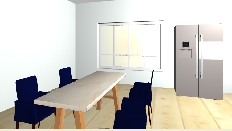}
    \end{subfigure}
    \caption*{(a) Input image from camera}
    \begin{subfigure}[b]{0.3\textwidth}
    \centering
    \includegraphics[width=\textwidth]{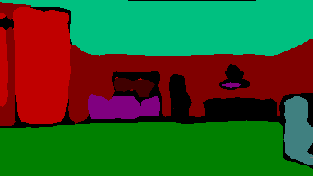}
    \end{subfigure}
    \hspace{0.1 cm}
    \begin{subfigure}[b]{0.3\textwidth} 
    \centering
    \includegraphics[width=\textwidth]{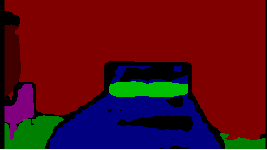}
    \end{subfigure}
    \hspace{0.1 cm}
    \begin{subfigure}[b]{0.3\textwidth}
    \centering
    \includegraphics[width=\textwidth]{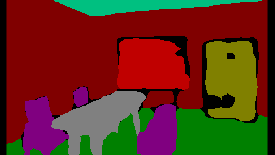}
    \end{subfigure}
    \caption*{(b) Semantic segmentation from input images}
    \begin{subfigure}[b]{0.3\textwidth}
    \centering
    \includegraphics[width=\textwidth]{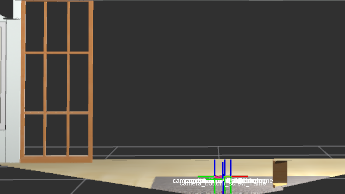}
    \end{subfigure}
    \hspace{0.1 cm}
    \begin{subfigure}[b]{0.3\textwidth} 
    \centering
    \includegraphics[width=\textwidth]{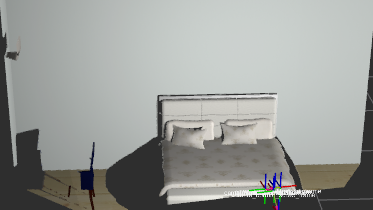}
    \end{subfigure}
    \hspace{0.1 cm}
    \begin{subfigure}[b]{0.3\textwidth}
    \centering
    \includegraphics[width=\textwidth]{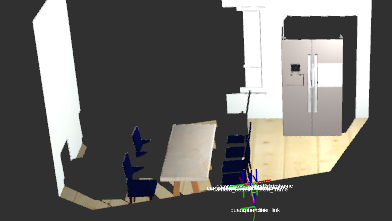}
    \end{subfigure}
    \caption*{(c) Color point cloud from input images}
    \begin{subfigure}[b]{0.3\textwidth}
    \centering
    \includegraphics[width=\textwidth]{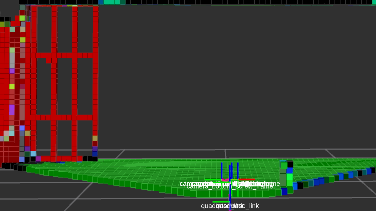}
    \end{subfigure}
    \hspace{0.1 cm}
    \begin{subfigure}[b]{0.3\textwidth} 
    \centering
    \includegraphics[width=\textwidth]{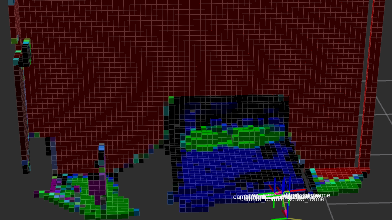}
    \end{subfigure}
    \hspace{0.1 cm}
    \begin{subfigure}[b]{0.3\textwidth}
    \centering
    \includegraphics[width=\textwidth]{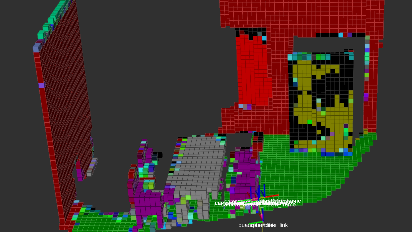}
    \end{subfigure}
    \caption*{(d) 3D semantic mapping from input images}
    \begin{subfigure}[b]{\textwidth}
    \centering
    \includegraphics[width=0.935\textwidth]{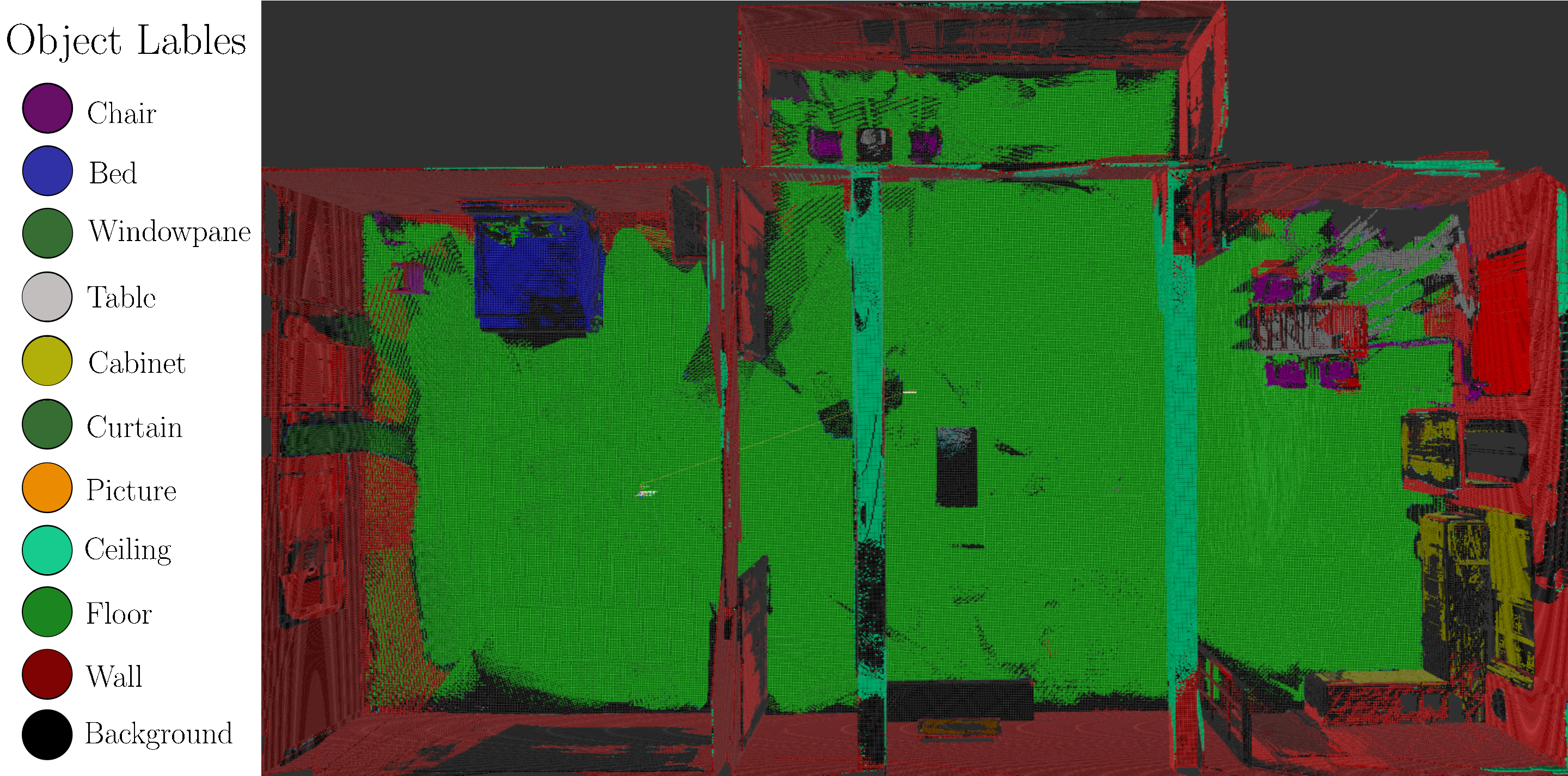}
    \end{subfigure}
    \caption*{(e) Overall 3D semantic mapping}
    \caption{3D visual representation of the obtained semantic maps}
    \label{fig:semantic_map}
\end{figure*}

Fig.~\ref{fig:semantic_map} shows the sequential processing stages and corresponding outcomes achieved by the proposed S3M system. As observed, our proposed method can integrate incoming semantic segmentation information (Fig. 9b) from input images (Fig. 9a) into the map volume, and the sparse map creation (Fig. 9d) process finalizes the OctoMap reconstruction by utilizing the point cloud containing semantic information. The implementation of the proposed system on the Jetson Xavier AGX platform, operating at 2Hz, where the object segmentation phase consumes 40ms per frame. The mapping results underscore the system's prowess in achieving real-time semantic mapping capabilities.

\section{Conclusion} \label{sec:conclusion}
In this paper, we introduced a novel approach for Semantic Sparse Mapping (S3M) in Unmanned Aerial Vehicles (UAVs) based on RGB-D camera data. Our proposed S3M SLAM framework addresses the challenge of integrating semantic information into UAV mapping operations, enabling enhanced perception and understanding of the environment. By fusing object instance segmentation with Octomap-based mapping, we achieve the creation of a semantic map that captures both spatial occupancy and object semantics. Future work could explore the integration of additional sensors to decrease cost and machine learning techniques to further enhance the UAV's perception capabilities.

% \subsection*{\textbf{Key References}}
\section*{Acknowledgment}
This work was supported by the Asian Office of Aerospace Research and Development under Grant/Cooperative Agreement Award No. FA2386-22-1-4042.

\bibliographystyle{IEEEtran}
\bibliography{ref}

\end{document}